\documentclass[10pt,letterpaper,onecolumn]{article}
\usepackage{amsmath,amssymb,graphicx,cite}
\usepackage{lettrine}
\usepackage[noend]{algpseudocode}
\usepackage{algorithmicx,algorithm}
\begin{document}% End of preamble and beginning of text.
\title{\textsf{\textbf{Efficient phase retrieval based on dark fringe recognition with an ability of bypassing invalid fringes}}}% Declares the document's title.

\author{\textsf{\textbf{Wen-Kai Yu}}\renewcommand{\thefootnote}{\arabic{footnote}}\footnotemark[1] \renewcommand{\thefootnote}{\fnsymbol{footnote}}\footnotemark[1],
\textsf{\textbf{An-Dong Xiong}}\renewcommand{\thefootnote}{\arabic{footnote}}\footnotemark[1],\\
\textsf{\textbf{Xu-Ri Yao}}\renewcommand{\thefootnote}{\arabic{footnote}}\footnotemark[2],
\textsf{\textbf{Guang-Jie Zhai}}\renewcommand{\thefootnote}{\arabic{footnote}}\footnotemark[2],
\textsf{\textbf{and Qing Zhao}}\renewcommand{\thefootnote}{\arabic{footnote}}\footnotemark[1]}      % Declares the author's name.
\renewcommand{\thefootnote}{\arabic{footnote}}
\footnotetext[1]{Center for Quantum Technology Research, School of Physics, Beijing Institute of Technology, Beijing 100081, China}
\footnotetext[2]{Key Laboratory of Electronics and Information Technology for Space System, National Space Science Center, Chinese Academy of Sciences, Beijing 100190, China}
\renewcommand{\thefootnote}{\fnsymbol{footnote}}
\footnotetext[1]{yuwenkai@bit.edu.cn}
\date{}% Deleting this command produces today's date.

\maketitle% Produces the title.

%\begin{abstract}
\noindent\textsf{\textbf{This paper discusses the noisy phase retrieval problem: recovering a complex image signal with independent noise from quadratic measurements. Inspired by the dark fringes shown in the measured images of the array detector, a novel phase retrieval approach is proposed and demonstrated both theoretically and experimentally to recognize the dark fringes and bypass the invalid fringes. A more accurate relative phase ratio between arbitrary two pixels is achieved by calculating the multiplicative ratios (or the sum of phase difference) on the path between them. Then the object phase image can be reconstructed precisely. Our approach is a good choice for retrieving high-quality phase images from noisy signals and has many potential applications in the fields such as X-ray crystallography, diffractive imaging, and so on.}}
%\end{abstract}

\section*{\textsf{Introduction}}% \subsection and \subsubsection commands.
\lettrine[lines=2]{I}{n} many optical measurement systems, detection devices can only record the Fourier magnitude-square (intensity) of the signal and are not allowed to directly measure the phase, because they can not catch the electromagnetic field which oscillates at frequency of $\sim10^{15}$ Hz. It is well known that the Fourier phase is often more important than the Fourier magnitude in recovering the structural information of the object \cite{Oppenheim1981}. Determining the unknown phase of a complex-valued function from given magnitude measurements of its Fourier transform, also known as the phase retrieval (PR) problem \cite{Patterson1934,Patterson1944}, has numerous applications, including X-ray crystallography \cite{Millane1990,Harrison1993}, diffractive imaging \cite{Miao1999,Bunk2007}, biomedical imaging \cite{Song2008}, astronomical imaging \cite{Dainty1987}, Fourier ptychography \cite{Zheng2013}, and digital holography \cite{Jueptner2005}, among others. To the best of our knowledge, the most widely used methods are based on the alternating projections (APs) between the time-domain/real-space and the Fourier magnitude constraints \cite{Gerchberg1972,Fienup1978,Fienup1982,Bauschke2003}, but rely heavily on the priori information about the signals \cite{Hayes1982,Fienup1983} and do not come with global convergence guarantees. Additionally, there exists a significant computational burden since one has to compute fast Fourier transforms at each iteration. Recently, there appeared new approaches based on convex optimization that use matrix-lifting to recast PR as a semi-definite programming (SDP) problem \cite{Chai2011,Voroninski2013}. The typical algorithms include PhaseLift \cite{Voroninski2013,Candes2013}, Wirtinger flow (WF) \cite{Candes2014}, truncated WF \cite{Chen2015,Kolte2016}, sparsity based PR \cite{Newton2012,Shechtman2011}. Generally, these methods are guaranteed to yield global solutions, utilizing a few coded diffraction patterns (CDPs) of a spatial light modulator (SLM), with a pattern number on the order of $(\log n)^4$ ($n$ is the length of the signal) \cite{Soltanolkotabi2013}. However, most of them are sensitive to noise and the number of patterns can still be reduced. Besides, they become computationally prohibitive as the dimension of the signal increases. Furthermore, since the coefficient values in the center of the Fourier plane are in orders of magnitude higher than those around, it is hence hard for a photodetector to record correctly the magnitude of both the high-frequency (detail) and low-frequency (contour) coefficients, resulting in the inaccuracy of the final phase retrieval.

In this work, instead of using AP or convex optimization, we follow a different route by adopting the dark fringe judgement and path searching to address the PR problem. That is, extract the dark fringes from the measured magnitude images on the image plane (rather than the Fourier/focal plane); then logically judge whether there exists a phase difference between adjacent pixels; next plan the path between two arbitrary pixels in order to bypass the invalid fringes caused by the measurement noise; and finally retrieve the complex image of the object by computing the multiplicative ratios on the path. We theoretically explain the mechanism of the dark fringe between two adjacent pixel-regions and analyze the relationship between the dark fringes and the point spread function. A series of experiments are conducted by placing a phase-modulation element in a laser-illuminated setup. The experimental results proved that the object phase image can be reconstructed precisely with fewer magnitude measurements. Moreover, our imaging protocol also reduces the computational burden and outperforms other state-of-the-art approaches in the scenarios involving various noise, avoiding the problem of the Fourier magnitude measurements.

\section*{\textsf{Results}}
\textbf{Experiment.}
In our experimental setup, the SLM is illuminated by the collimated HeNe laser polarized coherent light beam, and the reflected light of the SLM is then sampled by a charge coupled device (CCD) camera via an imaging lens, i.e., the far-field diffraction intensity image is measured. Thus the problem is to recover the real and imaginary part of the object from the measured far-field intensity images.
\begin{figure}[htbp]
\centering\includegraphics[width=12cm]{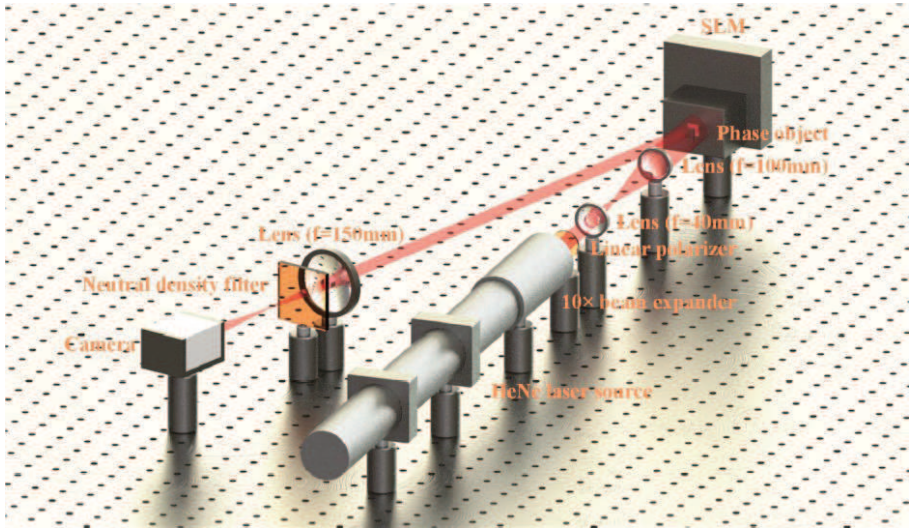}
\caption{Schematic of the experimental setup. The polarized coherent light from the HeNe laser illuminates the object and the SLM, and the emitted light is collected by an imaging lens of focal length 150~mm and of diameter 50.8~mm. Then several two-dimensional diffraction intensity images are recorded by a CCD camera. A $10\times$ complete beam expander (Edmund 55-578) with $\lambda/2$ MgF$_2$ coating and a combination of two lenses of focal length 40~mm and 100~mm respectively are used to make a total $25\times$ magnification. A 2 inch $\times$ 2 inch neutral density filter (Thorlabs NE220B) of optical densities (OD) 2 is used for the light attenuation, and the transmissivity is $10^{-\textrm{OD}}=0.01$.}
\label{fig:setup}
\end{figure}

Figure~\ref{fig:setup} presents a schematic of the experimental setup. A JDSU HeNe laser source (Model 1125) was used as the light source, with 5~mW random polarization and a beam diameter of 0.81~mm. In order to generate the polarized light required by the SLM, a {\O}25.0~mm linear polarizer (Thorlabs LPVISB100-MP2) is installed in an indexing rotation mount (Thorlabs RSP1X15/M), in front of the beam expander. The applied liquid crystal on silicon (LCoS) SLM (Meadowlark Optics HSPDM512-0635-PCIe-8-bit) has $512\times512$ (262,144) active pixels, which are divided into $16\times16$ (256) equally sized square segments of per-size $32\times32$ pixels in the optimization process. The phase stroke of the SLM is $2\pi$ at 635~nm, and the pixel pitch is $15\ \mu$m$\times\ 15\ \mu$m. The maximum diffraction efficiency is 90$\thicksim$95\% at 635~nm, and the switching frequency is larger than 60~Hz. For imaging, the measurement time for each frame was 38.5~ms. All images were obtained with a Point Grey GS3-U3-28S4M-C camera whose senor is a Sony ICX687 mono CCD, 1/1.8", with the maximum resolution of 1928$\times$1448, and pixel size of $3.69\ \mu$m$\times\ 3.69\ \mu$m. The maximum signal to noise ratio (SNR) of the camera is 39.56~dB.

\begin{figure}[htbp]
\centering\includegraphics[width=12cm]{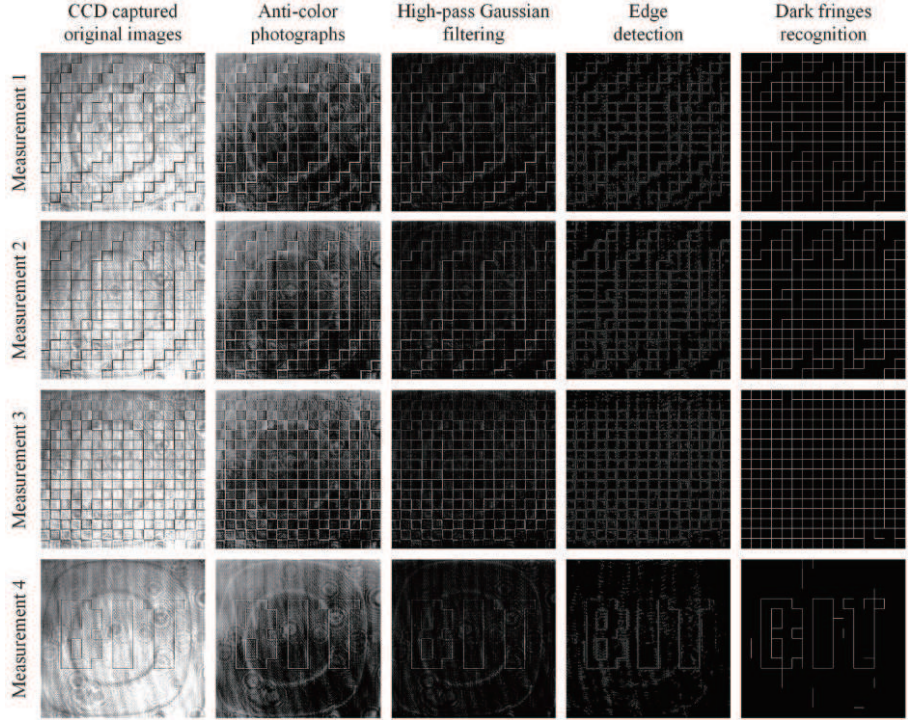}
\caption{Dark fringes extracted from the images recorded on the CCD. From the first column to the fifth column, we present the CCD captured original images, the anti-color photographs, the images that are generated by the high-pass Gaussian filtering, the edge detection images, and the images of the dark fringes recognition, respectively. Rows 1-4 stand for the first, the second, the third and the fourth measurement, according to four different phase modulation patterns.}
\label{fig:linejudgement}
\end{figure}

\textbf{Experimental results.}
From the measured images of the camera on the image plane (rather than the Fourier/focal plane, thus it avoids the problem of the Fourier magnitude measurements), we find that there may appear some dark fringes between adjacent pixel-units, if the size of single pixel-unit is large enough. Based on this phenomenon, we can extract the dark fringes from the measured images, using some image pre-processing technologies such as color inversing, high-pass Gaussian filtering, edge detection, and dark fringe recognition. An experimental example is shown in Figure~\ref{fig:linejudgement}. Then the phase difference between each pair of adjacent pixels can be calculated. Finally, the complex image of the original object can be reconstructed by computing the multiplicative ratios of the phase difference on the planned path between two arbitrary pixels. The detail of our protocol is presented in the section of Methods.

In our experiment, we set the resolution of phase modulation patterns to be $32\times32$ pixel size, which is also the minimum resolving size of the object. We here encode four phase patterns (actually $512\times 512$ pixels for each frame and can be easily converted to $16\times16$ pixels) on the SLM. The reasons for applying such big pixel-units are listed as follows: i) to reduce the possibility of the misjudgment caused by the fine grids (which are mainly the interference result of the light from the pixel gaps of the SLM); ii) to improve the flux of each pixel-unit and decrease the influence of noise, and thus improve the signal-to-noise ratio; iii) to obtain wider dark fringes that consist of more pixels of the CCD. The four patterns chosen in our experiment satisfy the condition: $M_{c+1,d+1}^j=q_j^*M_{c,d+1}^j=q_j^*M_{c+1,d}^j$, $q_j^*=i,-i,-1,1$, $j=1,2,3,4$, $c$ and $d$ are the $x$-$y$ coordinates, i.e., the phase difference between every two adjacent pixel-units is the same for each phase pattern. Since $q_j^*$ also equals to $e^{i\theta}$, where $q_j^*=\frac{1}{2}\pi,\frac{3}{2}\pi,\pi,2\pi$, corresponding to 63, 191, 127 and 0 of the patterns loaded on the SLM, which is in 8 bits precision. It is particularly worth mentioning here that the dark fringes will disappear when the phase difference in the same pixel-unit positions of one pattern and the original object image are conjugate. In computation, since the flux mainly focuses on the center of the CCD, we cut off the upper and lower edges (a few rows) of original images of the CCD.
\begin{figure}[htbp]
\centering\includegraphics[width=12cm]{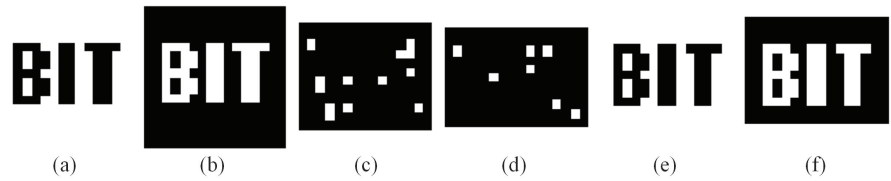}
\caption{Reconstructed results. (a) and (b) are the real part and imaginary part of the digital object, respectively; The row invalid boundary matrix $Matrix_a$ (c) and the column invalid boundary matrix $Matrix_b$ (d) show the locations of the misjudged row lines and the misjudged column lines; and the generation method of these two matrices can be found in Algorithm S1 of the Supplementary Information. (e) and (f) are the real part and imaginary part of the retrieved images, respectively.}
\label{fig:reconstruction}
\end{figure}

Although we have tried to get a better picture of the dark fringes, there still exists some misjudgment, as showed in Figures~\ref{fig:reconstruction}(c)--(d). The detail is described in Sec. Methods. That is why we design a path searching algorithm in order to bypass these misjudged lines. Then we calculate the relative ratio of each pixel according to the path planned. Finally, the real part and the imaginary part of the reconstructed result can be obtained very precisely, as shown in Figures~\ref{fig:reconstruction}(e)--(f).

\textbf{Dark fringes caused by the phase difference.}
Now let us analyze the relationship between the dark fringes and the point spread function (PSF). It is well known that the PSF describes the response of an imaging system to a point source or a point object. We can treat the PSF as the impulse response of a focused optical system. In a non-coherent imaging system, the image formation process is linear in power, while in a coherent system, the imaging is linear in the complex field. In our experiment scheme, each pixel of the light field reflected from the SLM can be seen as a point source, and the image plane can be regarded as the end. For simplicity, we assume that there is no scaling between the source and the object, and that the PSF of each point source is the same. Since there is a one-to-one correspondence between each point source on SLM and each pixel on the image plane, the PSF has a good translation consistency. We thus have $PSF(x,y)_{(a,b)}=PSF(x-a,x-b)_{(0,0)}$, where $(a,b)$ and $(x,y)$ denote the coordinates on the source plane and the imagine plane, respectively. We can assume $PSF(x,y)_{(0,0)}=p(x,y)c(x,y)$, where $p(x,y)$ is a positive real function and $c(x,y)$ is a complex function with unit norm and can be set to be $c(x,y)=1$ for convenience. The function $p$ is both differentiable and integrable, and the PSF is both axisymmetric and centrosymetric. Furthermore, it is required that the pixel size on the source plane should be larger than the radius of the PSF, thus only the interference of adjacent pixels needs to be considered. Due to the fact that the pixel shape is generally a rectangle, it is of good symmetry. For the fixed $y$ coordinate, we assume the primitive function of $p(x)$ is $P(x)$, and set $P(0)=0$, then we get $P(-x)=-P(x)$. The derivative of $p(x)$ is denoted as $p'$, then $p'(x)=-p'(-x)$. Assume that there are two horizontal adjacent pixels with the same amplitude, and phases $\varphi_1$, $\varphi_2$ corresponding to the left and the right pixel, respectively. The four boundary $x$-coordinates of the adjacent pixels are denoted as $a_1$, $a_2$ for the left, and $a_3$ and $a_4$ for the right, respectively. As mentioned above, the PSF has the translation consistency, thus we can move the coordinate origin to the symmetry $y$-axis of these two adjacent pixels. Consequently, the light intensity that casts on the image plane can be written as:
\begin{eqnarray}
\Gamma &=|e^{i\varphi_1}\int_{a_1}^{a_2}p(x-x_1)dx_1+e^{i\varphi_2}\int_{a_3}^{a_4}p(x-x_2)dx_2|^2
\end{eqnarray}
where $x_1$, $x_2$ are the independent variables. We then calculate the first and second derivatives of $\Gamma$ (see the Supplementary Information) in the case of $a_1=-a_4$ (the adjacent pixels are in the same length, actually, $a_1$ does not need to strictly equal to $a_4$, as long as both of them are much larger than the radius $r$ of the PSF, the following proof remains valid), $a_2=0$ (on any fixed $y$ coordinate axis): $\Gamma'=0$ and
\begin{eqnarray}
\Gamma''&=4[(1-\cos(\varphi_1-\varphi_2))(p(a_1)-p(0))^2\notag\label{eq:gramma1}\\
&+(1+\cos(\varphi_1-\varphi_2))P(a_1)p'(a_1)].
\end{eqnarray}

As proved in the Supplementary Information, the second term can be ignored when the locality of the PSF is well satisfied, thus Eq.~\ref{eq:gramma1} can be simplified to $\Gamma''=4[(1-\cos(\varphi_1-\varphi_2))(p(a_1)-p(0))^2]$. Since this function is always greater than or equal to 0, we can get a minimum that appears on the symmetry axis in the form of a dark fringe. The same goes for the analysis on the $x$ symmetry axis. The sharpness of the dark fringes depends on the phase difference between these two adjacent pixels. The complete theoretical demonstration is presented in the Supplementary Information.

\begin{figure}[htbp]
\centering\includegraphics[width=12cm]{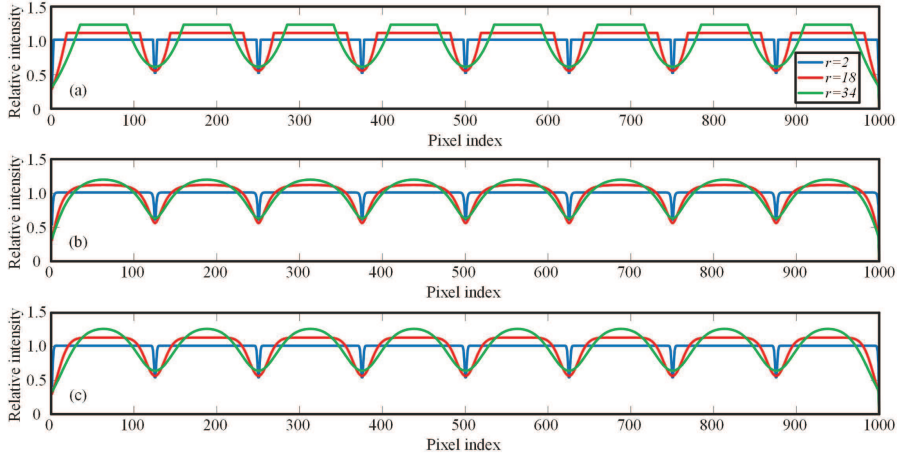}
\caption{The simulation results of the relative intensity of a one-dimensional phase signal $e^{i\omega\pi}$, of length 1000 pixels, under three kinds of PSFs. (a)--(c) show the results of $PSF_1$, $PSF_2$ and $PSF_3$, respectively. From the comparison, we can see that they have almost the same relative intensity at the location of the dark fringes.}
\label{fig:PSF}
\end{figure}

\begin{figure}[htbp]
\centering\includegraphics[width=10cm]{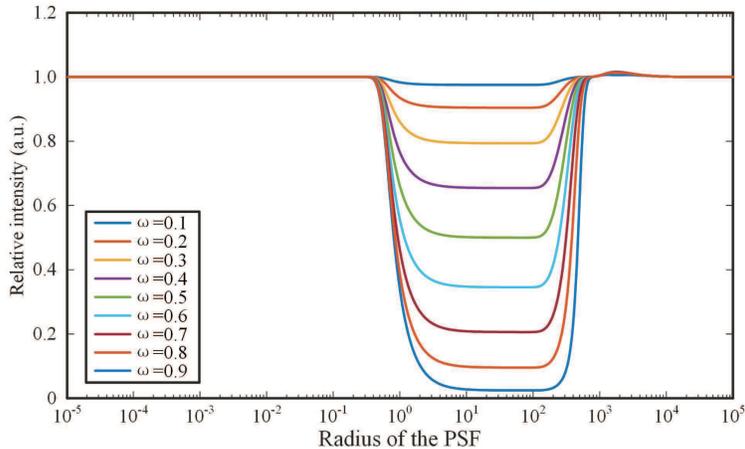}
\caption{Relative intensity (a.u.) of the dark fringes v.s. the radius of the PSFs. The length of each phase pixel-unit has 512 pixels. The PSF used here equals to $e^{-{(\frac{x}{r})}^2}$. The phase difference is from $0.1\pi$ to $0.9\pi$. When the PSF is over-spreading, the intensity of the symmetry axis will become a maximum.}
\label{fig:PSFradius}
\end{figure}

In order to verify the correctness of the above formulas, some simulations were made, as shown in Figures~\ref{fig:PSF}--\ref{fig:PSFradius}. The former was performed under different regions of integration of the corresponding multiple axes of symmetry, while the latter was done under the same region of integration of the center axis of symmetry but with the changing radius of the PSF as the $x$-coordinate (thus the interaction effects of multiple pixel-units had been characterized).

It is demonstrated that as long as the PSF satisfies the above restriction, the relative intensity (a.u.) of the dark fringes is independent of the kind of the PSF, but is relevant to the radius of the PSF. Figure~\ref{fig:PSF} shows the relative intensity of the light field on the image plane, for the same one-dimensional phase vector $e^{i\Omega\pi}$, where
\begin{eqnarray}
\Omega=&[\underbrace{0.5,\cdots,0.5}_{125},\underbrace{0,\cdots,0}_{125},\underbrace {-0.5,\cdots,-0.5}_{125},\underbrace{1,\cdots,1}_{125},\notag\\
&\underbrace{0.5,\cdots,0.5}_{125},\underbrace{0,\cdots,0}_{125},\underbrace {-0.5,\cdots,-0.5}_{125},\underbrace{1,\cdots,1}_{125}],
\end{eqnarray}
under different kinds of PSFs. Here, $PSF_1=\left\{ {\begin{array}{*{20}{c}}
{1{\rm{ }}\left| x \right| \le r}\\
{0{\rm{ }}\left| x \right| > r}
\end{array}}\right.$, $PSF_2=e^{(-\frac{{2\left|x\right|}}{r})}$, $PSF_3=e^{-{(\frac{x}{r})}^2}$, where the radius $r$ belongs to the pixel-set $\{2,18,34\}$. We see that the relative intensity of the dark fringes shows almost similar behaviors no matter what kind of the PSF is used.

The steep peak presented in Figure~\ref{fig:PSFradius} is in accordance with Eq.~\ref{eq:gramma1}. Here, the PSF applied is $e^{-{(\frac{x}{r})}^2}$, and the one-dimensional phase vector is $e^{i\Omega\pi}$, where $\Omega=[\omega,0,\omega,0,\omega,0,\omega,0]$, $\omega=0.1,0.2,\cdots,0.9$. Hence there are 8 pixel-units in total, each of which contains 512 pixels. We use the center position as the axis of symmetry, and change the radius $r$ of the PSF as well as the phase difference. When the effective region of integration is larger than two pixel-units, the influence of other pixel-units must be considered, which makes the problem more complex. As mentioned above, the first term of Eq.~\ref{eq:gramma1} is always nonnegative, and the second term $(1+\cos(\varphi_1-\varphi_2))P(a_1)p'(a_1)$ can be ignored when the locality of the PSF is well satisfied, i.e., $a_i>>r$, $i=1,2,3,4$ (see the Supplementary Information). However, when the locality is poor or the PSF is over-diffusive, this second term cannot be neglected and will always be negative (the PSF map expresses the descended distribution trend from the center to the periphery, thus if $a>0$, $P(a_1)$ is positive, $p'(a_1)$ is negative; otherwise, if $P(a_1)$ is negative, $p'(a_1)$ is positive), especially for the small phase difference, making the axis of symmetry be the maximum position. In other words, if the locality exceeds the boundary of pixel-unit, it will produce the negative interaction effects of a plurality of pixel-units, which cannot be neglected. From Figure~\ref{fig:PSFradius}, it can also be seen that the left parts of the curves is of high relative intensity, for the reason that the radius of PSF here is less than the length of one pixel-unit, and the dark fringes can not be distinguished by a CCD. Actually, in order to detect the dark fringes, the width of them should be greater than at least one-pixel-width of the CCD. It is important to note that the value range of the radius is wide according to the low-lying areas presented in the curves.

In a nutshell, the selection of the PSF in our optical system is flexible with fault-tolerant guarantee. Moreover, in most cases, the accuracy of the measurements of the dark fringes depends on the phase difference to be distinguished.

\section*{\textsf{Summary and future prospects}}
Since phase retrieval plays an important role in various fields of science and engineering, including X-ray crystallography, diffractive imaging, electron microscopy, astronomical observation, and optical imaging, we believe that our approach is a much better choice for retrieving high-quality phase images from noisy signals with fewer magnitude measurements and less computational complexity, and can be used to improve existing phase retrieval schemes and applications, with rapid future development of computer vision algorithms. In the future, we will continue to experimentally realize phase imaging of ultra-weak three-dimensional objects by exploiting the sparsity of the signals, for applications such as X-ray crystallography. An extended experimental setup, some quantitative analysis of the noise as well as the influence factors of the dark fringes will be presented in a future paper.

\section*{\textsf{Methods}}
\textbf{Protocol.}
In the practical optical imaging system, for the fact that it is far easier for the detectors to record the modulus of diffraction pattern than to measure its phase, we only have access to the magnitude measurements. Traditionally, suppose $x\in\mathbb{C}^n$ is a discrete signal and we are given the information about the squared modulus of the inner product between $m$ known sampling vectors $z_j$ and the signal $x$, namely, $b_j=\left|\langle{x,z_j}\rangle\right|^2,\ j=1,...,m$, where $b_j$ are the observed measurements. This problem is generally an instance of a quadratic program. Instead of using the traditional methods like AP, SDP etc., here we introduce a new approach to recover the missing phase of the data by essentially solving a dark fringe judgement (logical judgment) problem.

At first, we implement image pre-processing on the measured images, including color inversing, high-pass Gaussian filtering, edge detection, dark fringe recognition, then we get multiple images of identified dark fringes. A example of the specific generation method is given in the Supplementary Information.

We only need the phase difference in the same position of the pattern and the original image to be conjugate. The detailed constructing method of patterns and reference library matrix can be found in Sec. S2 of the Supplementary Information.

If there exists a dark fringe between two adjacent pixel-units, then they must have a phase difference. Each modulation pattern $a_j$ of $s_1\times s_2$ pixel-size has a geometric proportion $q_j^*$, $j=1,...,m$. In our protocol, since the number of modulation patterns is only related to the number of the phase value quantization, rather than the length of the signal, the magnitude measurements needed for the precise phase image reconstruction is much fewer than traditional methods. It is easy to determine the whether or not there exists a phase difference between each two adjacent pixel-units for all images, saved as $Row_j$ of pixel-size $s_1\times (s_2-1)$ and $Col_j$ of pixel-size $(s_1-1)\times s_2$, where $j=1,2,3,\ldots,m$. However, because of the measurement noise, there may exist some unconformity between the dark fringes and the original relative ratios. The system noise includes the intrinsic noise of the CCD, the Newton's rings caused by multiple surfaces of optical devices, as well as the fine grids generated by the interference of the light reflected from the substrate and the liquid crystal pixels of the SLM. All of these factors make it much more difficult for the computer to resolve the dark fringes. Consequently, it is necessary to extract the misjudged ones from the dark fringes. For the fact that the disappearance of a dark fringe only happens in one measurement of these modulation patterns, the dark fringe which does not disappear in any pattern modulation process or disappears in more than once modulation should be marked as a misjudged line. In our main algorithm (see the Supplementary Information: Algorithm S1), we just need to determine whether $\sum\limits_{j=1}^m{Row_j}$ and $\sum\limits_{j=1}^m{Col_j}$ is equal to $m-1$, and if not, the indices of such elements will be recorded and these elements are undoubtedly the misjudged lines. To mark out these misjudged lines, a row invalid boundary matrix $Matrix_a$ and a column invalid boundary matrix $Matrix_b$ are produced. A new path will be designed automatically via path searching algorithm to bypass these misjudged lines. The main design idea of our path searching algorithm (see the Supplementary Information: Algorithm S2) is that we firstly plan up or down paths for any two points those are in the same column and without the invalid boundary line between them; then plan the paths of turning right or left for any two points which are located in adjacent columns, with the help of a feasible point in the column of the origin point; the last and the most important searching step for the two points with the column distance greater than one, is realised by making full use of the relay of feasible paths, i.e., the path from origin to an intermediate point/pixel is connected with the path from the intermediate point/pixel to the destination.

\noindent\textbf{Discussion.}
In fact, some invalid row boundary lines and invalid column boundary lines may form some semi-closed shapes with the opening direction contrary to the flowing direction of the paths. Figuratively speaking, we compare these semi-closed shapes to the umbrella skeleton of an umbrella, then the path planning looks like the rain hitting on the umbrella, but not the people under the umbrella, which will finally result in a dead end. In order to avoid this problem, we transpose and exchange $Matrix_a$ and $Matrix_b$, and compute the paths again. Assume the appearing probability of misjudged lines is $\sigma<<1$, if we perform path searching algorithm for once, the of occurrence probability of a path blocking event is $O(\sigma^3(1-\sigma))$. With the matrix $Matrix_a$ and $Matrix_b$ transposed and exchanged, the probability of a path blocking event happens can drop to $O({\sigma^4})$, with some negligible computation cost (only a few seconds). Additionally, we can also change the starting point to solve the problem of semi-closed shapes. Another thing worth noting is that a fully enclosed shape of the boundary lines must lead to a dead end. But the occurrence probability of such an event is $O({\sigma^4})$, which is low enough to be ignored.

Furthermore, the path searching algorithm used here is designed especially for planning path between multiple starts and ends, enabling us to get an average ratio result by starting from different start points. Therefore, it will be important for getting better phase resolution and accuracy.

After the paths of every pixels have been planned, according to the reference library matrix $R$, we can acquire a more accurate relative phase ratio between two arbitrary pixels by calculating the multiplicative ratios of $q_j$ (or the sum of phase difference) on the path between them, avoiding the negative influence of misjudged lines. Then the object phase image can be reconstructed precisely. Since our imaging protocol only need to make the dark fringe judgement (logical judgment) and the path searching, it also reduces the computational burden.

\vspace{1em}\noindent\textsf{\textbf{Received X XX 2016}}%; accepted X XX XXXX;\\ published online X XX XXXX}}

\section*{\textsf{Acknowledgments}}
\textbf{This} work was supported by the National Key Scientific Instrument and Equipment Development Project of China (Grant No. 2013YQ030595), the National High Technology Research and Development Program of China (Grant No. 2013AA122902), National Natural Science Foundation of China (11675014). We are grateful to Ning Wu and Da-Zhi Xu for offering direct help in improving the manuscript and refining the idea. We warmly acknowledge Fei-Long Wang for preparing Figure~\ref{fig:setup}. The authors also thank the help of Jing-Yu Liu during the early stages of this work.

\section*{\textsf{Author Contributions}}
Wen-Kai Yu proposed the idea, designed the experimental setup and performed all experiments, analysed all data and drafted the manuscript. An-Dong Xiong performed the theoretical simulation, assisted the experiments as well as the writing of the manuscript. Xu-Ri Yao provided some valuable advices on the experiments. Guang-Jie Zhai and Qing Zhao supervised the research.

\section*{\textsf{Additional information}}
Supplementary information is available in the online version of this paper. Reprints and permissions information is available online at the journal website. Correspondence and requests for materials should be addressed to Wen-Kai Yu.

\section*{\textsf{Additional information}}
The authors declare no competing financial interests.
\end{document}